\documentclass[letterpaper, 12pt]{article}
\usepackage[margin=2.75cm]{geometry}
\usepackage[affil-it]{authblk}

\makeatletter
\def\@maketitle{%
	\newpage
	\null
	\vskip 2em%
	\begin{center}%
		\let \footnote \thanks
		{\Large\bfseries \@title \par}%
		\vskip 1.5em%
		{\normalsize
			\lineskip .5em%
			\begin{tabular}[t]{c}%
				\@author
			\end{tabular}\par}%
		\vskip 1em%
		{\normalsize \@date}%
	\end{center}%
	\par
	\vskip 1.5em}
\makeatother

\providecommand{\keywords}[1]{\textbf{\textit{Keywords: }} #1}

\usepackage{times}

\usepackage{epstopdf}
\usepackage{mathtools}
\usepackage{amsmath,amsfonts,amssymb,amsthm}
\usepackage{graphicx}
\usepackage{subfig}
\usepackage[font=scriptsize,labelfont=bf,labelsep=space]{caption}
\usepackage{placeins}
\usepackage{float}
\usepackage{pgfplots}
\usepackage{enumerate}
\usepackage[utf8]{inputenc}
\usepackage[T1]{fontenc}
\usepackage{lmodern}
\usepackage{hyperref}
\usepackage{isomath}
\usepackage{xcolor}
\usepackage{algorithm}
\usepackage[algo2e]{algorithm2e}
\usepackage{algpseudocode}
\usepackage{xpatch}
\mathtoolsset{showonlyrefs}

\algnewcommand\algorithmicinput{\textbf{Input:}}
\algnewcommand\Input{\item[\algorithmicinput]}
\algnewcommand\algorithmicoutput{\textbf{Output:}}
\algnewcommand\Output{\item[\algorithmicoutput]}
\algnewcommand\algorithmicgiven{\textbf{Given}}
\algnewcommand\Given{\item[\algorithmicgiven]}
\algnewcommand\algorithmicchoose{\textbf{Choose}}
\algnewcommand\Choose{\item[\algorithmicchoose]}

\makeatletter
\xpatchcmd{\algorithmic}{\itemsep\z@}{\itemsep=0.8ex}{}{}
\makeatother

\newcommand{\R}{\mathbb R}
\newcommand{\N}{\mathbb N}
\newcommand{\PP}{\mathcal P}
\newcommand{\CC}{\mathcal C}
\renewcommand\vec[1]{\mathbf{#1}}

\usepackage{tikz}
\usetikzlibrary{shapes.geometric, shapes.multipart, shapes, arrows, shadings,fit,calc}
\tikzstyle{blank} = []
\tikzstyle{block} = [rectangle, rounded corners, minimum width=6em, minimum height=5em,text centered, draw=black]
\tikzstyle{line} = [draw, -latex']
\tikzstyle{arrow} = [->,>=stealth]
\tikzstyle{cloud} = [draw, ellipse, node distance=3cm,minimum height=2em]

\graphicspath{{figures/}, {tables/}}

\title{A deep learning approach to\\ diabetic blood glucose  prediction}
	\author[1]{H.N. Mhaskar}
	\affil[1]{Institute of Mathematical Sciences, Claremont Graduate University, Claremont, CA, USA}
	\author[2]{S.V. Pereverzyev}
	\affil[2]{Johann Radon Institute, Linz, Austria}
	\author[3]{M.D. van der Walt
		\thanks{Electronic address: \texttt{maryke.thom@gmail.com}}}
	\affil[3]{Department of Mathematics, Vanderbilt University, Nashville, TN 37240, USA}
	
	\date{}

\begin{document}

\maketitle

\begin{abstract}
We consider the question of 30-minute prediction of blood glucose levels measured by continuous glucose monitoring devices, using clinical data. While most studies of this nature deal with one patient at a time, we take a certain percentage of patients in the data set as training data, and test on the remainder of the patients; i.e., the machine need not re-calibrate on the new patients in the data set. We demonstrate how deep learning can outperform shallow networks in this example. One novelty is to demonstrate how a parsimonious deep representation can be constructed using domain knowledge.\\

\noindent \keywords{{deep learning, deep neural network, diffusion geometry, continuous glucose monitoring, blood glucose prediction.}} 
\end{abstract}

\section{Introduction}

\label{introsect}
Deep Neural Networks, especially of the convolutional type (DCNNs), have
started a revolution in the field of artificial intelligence and
machine learning, triggering a large number of commercial ventures and
practical applications. There is therefore a great deal of theoretical investigations
about when and why deep (hierarchical) networks perform so well compared to shallow ones. For example, Montufar and
Bengio \cite{montufar2014number} showed that the number of linear
regions that can be synthesized by a deep network with ReLU
nonlinearities is much larger than by a shallow network. Examples
of specific functions that cannot be represented efficiently by
shallow networks have been given very recently by Telgarsky
\cite{telgarsky2015representation} and by Safran and Shamir \cite{safran2016depth}. \\

It is argued in \cite{mhaskar2016deep} that from a function approximation point of view, deep networks are able to overcome the so-called curse of dimensionality if the target function is hierarchical in nature; e.g., a target function of the form
\begin{equation}
h_l(h_3(h_{21} (h_{11}(x_1, x_2), h_{12}(x_3, x_4)),\\
h_{22}(h_{13}(x_5, x_6), h_{14}(x_7, x_8)))),
\label{compositionfigure}
\end{equation}
where each function has a bounded gradient, can be approximated by a deep network comprising $n$ units, organized as a binary tree, up to an accuracy $\mathcal{O}(n^{-1/2})$. In contrast, a shallow network that cannot take into account this hierarchical structure can yield an accuracy of at most $\mathcal{O}(n^{-1/8})$. In theory, if samples of the functions $h, h_{1,2}, \ldots$ are known, one can construct the networks explicitly, without using any traditional learning algorithms such as back propagation.\\

One way to think of the function in \eqref{compositionfigure} is to think of the inner functions as the features of the data, obtained in a hierarchical manner.
While classical machine learning with shallow neural networks requires that the relevant features of the raw data should be selected by using domain knowledge before the learning can start,
deep learning algorithms appear to select the right features automatically. However, it is typically not clear how to interpret these features. Indeed, from a mathematical point of view, it is easy to show that a structure such as (\ref{compositionfigure}) is not unique, so that the hierarchical features cannot be defined uniquely, except perhaps in some very special examples.  \\

In this paper, we  examine how a deep network can
be constructed in a parsimonious manner if we do allow domain knowledge to suggest the compositional structure of the target function as well as the values of the constituent functions. We study the problem of predicting, based on the past few readings of a continuous glucose monitoring (CGM) device, both the blood glucose (BG) level and the rate at which it would be changing 30 minutes after the last reading. From the point of view of diabetes management, a reliable solution to this problem is of great importance. If a patient has some warning that that his/her BG will rise or drop in the next half hour,  the patient can take certain precautionary measures to prevent this (e.g., administer an insulin injection or take an extra snack) \cite{naumova2012meta,snetselaar2009nutrition}.\\

Our approach is to first construct three networks based on whether a 5-minute prediction, using ideas in \cite{mhaskar2013filtered}, indicates the trend to be in the hypoglycemic (0--70 mg/dL), euglycemic (70--180 mg/dL), or hyperglycemic (180--450 mg/dL) range. We then train a ``judge'' network to get a final prediction based on the outputs of these three networks. Unlike the tacit assumption in the theory in \cite{mhaskar2016deep}, the readings and the outputs of the three constituent networks are not dense in a Euclidean cube. Therefore, we use diffusion geometry ideas in \cite{ehler2012locally} to train the networks in a manner analogous to manifold learning.\\

From the point of view of BG prediction, a novelty of our paper is the following. 
Most of the literature on this subject which we are familiar with does the prediction patient-by-patient; for example, by taking 30\% of the data for each patient to make the prediction for that patient. 
In contrast, we consider the entire data for 30\% of the patients as training data, and predict the BG level for the remaining 70\% of patients in the data set. 
Thus, our algorithm transfers the knowledge learned from one data set to another, although it does require the knowledge of both the data sets to work with. From this perspective, the algorithm has similarity with the meta-learning approach by \cite{naumova2012meta}, but in contrast to the latter, it does not require a meta-features selection.\\

We will explain the problem and the evaluation criterion in Section~\ref{problemsect}. 
Some prior work on this problem is reviewed briefly in Section~\ref{priorworksect}. 
The methodology and algorithm used in this paper are described in Section~\ref{methodsect}. 
The results are discussed in Section~\ref{resultsect}. 
The mathematical background behind the methods described in Section~\ref{methodsect} is summarized in Appendix~\ref{mnpsect} and Appendix~\ref{funcapproxsect}.

\section{Problem statement and evaluation}\label{problemsect}

We use a clinical data set provided by the DirectNet Central Laboratory \cite{DirecNet2005}, which lists BG levels of different patients taken at 5-minute intervals with the CGM device; i.e., for each patient $p$ in the patient set $P$, we are given a time series $\{s_p(t_j)\}$, where $s_p(t_j)$ denotes the BG level at time $t_j$. Our goal is to predict for each $j$, the level $s_p(t_{j+m})$, given readings $s_p(t_j),\ldots, s_p(t_{j-d+1})$ for appropriate values of $m$ and $d$.  
For a 30-minute prediction, $m=6$, and we took $d=7$ (a sampling horizon $t_{j} - t_{j-d+1}$ of 30 minutes has been suggested as the optimal one for BG prediction in \cite{mhaskar2013filtered,hayes2009algorithm}). \\

In this problem, numerical accuracy is not the central objective. To quantify the clinical accuracy of the considered predictors, we use the Prediction Error-Grid Analysis (PRED-EGA) \cite{sivananthan2011assessment}, which has been designed especially for glucose predictors and which, together with its predecessors and variations, is by now a standard metric in the blood glucose prediction problem (see, for example, \cite{naumova2012adaptive,naumova2012meta,pappada2011neural,reifman2007predictive}) This assessment methodology records reference BG estimates paired with the BG estimates predicted for the same moments, as well as reference BG directions and rates of change paired with the corresponding estimates predicted for the same moments. As a result, the PRED-EGA reports the numbers (in percent) of Accurate (Acc.), Benign (Benign) and Erroneous (Error) predictions in the hypoglycemic, euglycemic  and hyperglycemic ranges separately. This stratification is of great importance because consequences caused by a prediction error in the hypoglycemic range are very different from ones in the euglycemic or the hyperglycemic range.\\

\section{Prior work}\label{priorworksect}

Given the importance of the problem, many researchers have worked on it in several directions. Relevant to our work is the work using a linear predictor and work using supervised learning.\\

The linear predictor method estimates $s_p'(t_j)$ based on the previous $d$ readings, and then predicts
\begin{equation}
s_p(t_{j+m})\approx s_p(t_j)+(t_{j+m}-t_j)s_p'(t_j).
\label{eq-linpred}
\end{equation} 
Perhaps the most classical of these is the work  \cite{savitzky1964smoothing} by Savitzky and Golay, that proposes an approximation of $s_p'(t)$ by the derivative of a polynomial of least square fit to the data $(t_k,s_p(t_k))$, $k=j,\ldots,j-d+1$. The degree $n$ of the polynomial acts  as a regularization parameter.
However, in addition to the intrinsic ill--conditioning of numerical differentiation, the solution of the least square problem as posed above involves a system of linear equations with the Hilbert matrix of order $n$, which is notoriously ill--conditioned. 
Therefore, it is proposed in \cite{lu2013legendre} to use Legendre polynomials rather than the monomials as the basis for the space of polynomials of degree $n$.  
A procedure to choose $n$ is given in  \cite{lu2013legendre}, together with error bounds in terms of $n$ and the estimates on the noise level in the data, which are optimal up to a constant factor for the method  in the sense of the oracle inequality. 
A substantial improvement on this method was proposed in \cite{mhaskar2013filtered}, which we summarize in Appendix~\ref{mnpsect}. As far as we are aware, this is the state of the art in this approach in short term blood glucose prediction using linear prediction technology.\\

There exist several BG prediction algorithms in the literature that use a supervised learning approach. These can be divided into three main groups. \\

The first group of methods employ kernel-based regularization techniques to achieve prediction (for example, \cite{naumova2011extrapolation,naumova2012meta} and references therein), where Tikhonov regularization is used to find the best least square fit to the data $(t_k,s_p(t_k))$, $k=j,\ldots,j-d+1$, assuming the minimizer belongs to a reproducing kernel Hilbert space (RKHS). Of course, these methods are quite sensitive to the choice of kernel and regularization parameters. Therefore, the authors in \cite{naumova2011extrapolation,naumova2012meta} develop methods to choose both the kernel and regularization parameter adaptively, or through meta-learning (``learning to learn'') approaches.\\

The second group consists of artificial neural network models (such as \cite{pappada2008development, pappada2011neural}). In \cite{pappada2011neural}, for example, a feed-forward neural network is designed with eleven neurons in the input layer (corresponding to variables such as CGM data, the rate of change of glucose levels, meal intake and insulin dosage), and nine neurons with hyperbolic tangent transfer function in the hidden layer. The network was trained with the use of data from 17 patients and tested on data from 10 other patients for a 75-minute prediction, and evaluated using the classical Clarke Error-Grid Analysis (EGA) \cite{clarke1987evaluating}, which is a predecessor of the PRED-EGA assessment metric. Classical EGA differs from PRED-EGA in the sense that it only compares absolute BG concentrations with reference BG values (and not rates of change in BG concentrations as well). Although good results are achieved in the EGA grid in this paper, a limitation of the method is the large amount of additional input information necessary to design the model, as described above.\\

The third group consists of methods that utilize time-series techniques such as autoregressive (AR) models (for example, \cite{reifman2007predictive,sparacino2007glucose}). In \cite{reifman2007predictive}, a tenth-order AR model is developed, where the AR coefficients are determined through a regularized least square method. The model is trained patient-by-patient, typically using the first 30\% of the patient's BG measurements, for a 30-minute or 60-minute prediction. The method is tested on a time series containing glucose values measured every minute, and evaluation is again done through the classical EGA grid.
The authors in \cite{sparacino2007glucose} develop a first-order AR model, patient-by-patient, with time-varying AR coefficients determined through weighted least squares. Their method is tested on a time series containing glucose values measured every three minutes, and quantified using statistical metrics such as measuring the mean square of the errors. As noted in \cite{naumova2012meta}, these methods seem to be sensitive to gaps in the input data.\\

\section{Methodology in the current paper}\label{methodsect}

Our proposed method represents semi-supervised learning that follows an entirely different approach from those described above. It is not a classical statistics/optimization based approach; instead, it is based on function approximation on data defined manifolds, using diffusion polynomials. In this section, we describe our deep learning method, which consists of two layers, in details.\\

Given the time series $\left\{ s_p(t_j) \right\}$ of BG levels at time $t_j$ for each patient $p$ in the patient set $P$, where $t_{j}-t_{j-1} = 5$ minutes, we start by formatting the data into the form $\left\{ \left( \vec{x}_j, y_j \right) \right\}$, where
$$ \vec{x}_j = \left( s_p(t_{j-d+1}), \cdots, s_p(t_j) \right) \in \R^d \quad \textup{and} \quad y_j = s_p(t_{j+m}) \in \R, \quad \textup{for all patients } \ p \in P. $$
We will use the notation
$$ \PP := \left\{ \vec{x}_j = \left( s_p(t_{j-d+1}), \cdots, s_p(t_j) \right) : p\in P \right\}. $$

We also construct the diffusion matrix from $\PP$. This is done by normalizing the rows of the weight matrix $\mathcal{W}^{\varepsilon}_N$ in \eqref{eq-diffusion-matrix}, following the approach in \cite[pp.~33-34]{lafon2004diffusion}.\\

Having defined the input data $\PP$ and the corresponding diffusion matrix, our method proceeds as follows.

\subsection{First layer: Three networks in different clusters}\label{layer1subsect}

To start our first layer training, we form the training patient set $TP$ by randomly selecting (according to a uniform probability distribution) $M$\% of the patients in $P$. The training data are now defined to be all the data $(\vec{x}_j, y_j)$ corresponding to the patients in $TP$. We will use the notations
$$ \CC := \left\{ \vec{x}_j = \left( s_p(t_{j-d+1}), \cdots, s_p(t_j) \right) : p\in TP \right\} $$ and
$$ \CC^{\star} := \left\{ (\vec{x}_j, y_j) = \left( \left( s_p(t_{j-d+1}), \cdots, s_p(t_j) \right), s_p(t_{j+m}) \right) : p\in TP \right\}.$$

Next, we make a short term prediction $L_{\vec{x}_j}(t_{j+1})$ of the BG level $s_p(t_{j+1})$ after 5 minutes, for all the given measurements $\vec{x}_j \in \CC$, by applying the linear predictor method (summarized in Section \ref{priorworksect} and Appendix~\ref{mnpsect}). Based on these 5-minute predictions, we divide the measurements in $\CC$ into three clusters $\CC_o, \CC_e$ and $\CC_r$,
\begin{gather*}
\CC_o = \left\{ \vec{x}_j \in \CC: 0 \leq L_{\vec{x}_j}(t_{j+1}) \leq 70 \right\} \ \textup{(hyp\textbf{o}glycemia)}; \\
\CC_e = \left\{ \vec{x}_j \in \CC: 70 < L_{\vec{x}_j}(t_{j+1}) \leq 180 \right\} \ \textup{(\textbf{e}uglycemia)}; \\
\CC_r = \left\{ \vec{x}_j \in \CC: 180 < L_{\vec{x}_j}(t_{j+1}) \leq 450 \right\} \ \textup{(hype\textbf{r}glycemia)},
\end{gather*}
with
$$ \CC_{\ell}^{\star} = \left\{ (\vec{x}_j, y_j) : \vec{x}_j \in \CC_{\ell} \right\}, \quad \ell \in \left\{ o,e,r\right\}. $$

The motivation of this step is to gather more information concerning the training set to ensure more accurate predictions in each BG range -- as noted previously, consequences of prediction error in the separate BG ranges are very different.\\

In the sequel, let $S(\Gamma,\vec{x}_j)$ denote the result of the  method of Appendix~\ref{funcapproxsect} (that is, $S(\Gamma,\vec{x}_j)$ is defined by equation \eqref{eq-sigma}), used with training data $\Gamma$ and evaluated at a point $\vec{x}_j\in\PP$. After obtaining the three clusters $\CC_o^{\star}, \CC_e^{\star}$ and $\CC_r^{\star}$, we compute the three predictors
\begin{equation}
f_{o}(\vec{x}) := S(\CC_o^{\star}, \vec{x}), \quad f_{e}(\vec{x}) := S(\CC_e^{\star}, \vec{x}) \quad \textup{and} \quad f_{r}(\vec{x}) := S(\CC_r^{\star}, \vec{x}), \quad \textup{for all } \ \vec{x}\in\PP,
\label{eq-threenetworks}
\end{equation}
as well as the ``judge'' predictor, based on the entire training set $\CC^{\star}$,
$$ f_{J}(\vec{x}) := S(\CC^{\star}, \vec{x}), \quad \vec{x}\in\PP, $$
using the summability method of Appendix~\ref{funcapproxsect} (specifically, \eqref{eq-sigma}). We remark that, as discussed in \cite{mhaskar2010eignets}, our
approximations in \eqref{eq-sigma} can be computed as classical radial basis function networks, with exactly
as many neurons as the number of eigenfunctions used in \eqref{phidef}. (As mentioned in Appendix~\ref{funcapproxsect}, this number is determined to ensure that the system \eqref{quadrature} is still well conditioned.)\\

The motivation of this step is to decide which one of the three predictions ($f_o, f_e$ or $f_r$) is the best prediction for each datum $\vec{x} \in \PP$. Since we do not know in advance in which blood glucose range a particular datum will result, we need to use all of the training data for the judge predictor $f_J$ to choose the best prediction.

\subsection{Second layer (judge): Final output}\label{layer2subsect}

In the last training layer, a final output is produced based on which $f_{\ell}, \ \ell\in\left\{ o,e,r \right\}$ gives the best placement in the PRED-EGA grid, using $f_J$ as the reference value. The PRED-EGA grid is constructed by using comparisons of $f_o$ (resp., $f_e$ and $f_r$) with the reference value $f_J$ -- specifically, it involves comparing
$$f_o(\vec{x}_j) \ (\textup{resp., } f_e(\vec{x}_j) \ \textup{and} \ f_r(\vec{x}_j)) \quad \textup{with} \quad f_J(\vec{x}_j)$$
as well as the rates of change
\begin{multline}
\dfrac{f_o(\vec{x}_{j+1}) - f_o(\vec{x}_{j-1})}{2(t_{j+1} - t_{j-1})} \ \left(\textup{resp. }, \dfrac{f_e(\vec{x}_{j+1}) - f_e(\vec{x}_{j-1})}{2(t_{j+1} - t_{j-1})} \ \textup{and} \ \dfrac{f_r(\vec{x}_{j+1}) - f_r(\vec{x}_{j-1})}{2(t_{j+1} - t_{j-1})} \right) \\
\textup{with} \quad \dfrac{f_J(\vec{x}_{j+1}) - f_J(\vec{x}_{j-1})}{2(t_{j+1} - t_{j-1})},
\label{eq-rateofchange}
\end{multline}
for all $\vec{x}_j \in \PP$. Based on these comparisons, PRED-EGA classifies $f_o(\vec{x}_j)$ (resp., $f_e(\vec{x}_j)$ and $f_r(\vec{x}_j)$) as being Accurate, Benign or Erroneous. As our final output $f(\vec{x}_j)$ of the target function at ${\vec{x}_j}\in\PP$, we choose the one among $f_{\ell}(\vec{x}_j), \ \ell\in\left\{ o,e,r \right\}$ with the best classification, and if there is more than one achieving the best grid placement, we output the one among $f_{\ell}(\vec{x}_j), \ \ell\in\left\{ o,e,r \right\}$ that has value closest to $f_J(\vec{x}_j)$. For the first and last $\vec{x}_j$ for each patient $p\in P$ (for which the rate of change \eqref{eq-rateofchange} cannot be computed), we use the one among $f_{\ell}(\vec{x}_j), \ \ell\in\left\{ o,e,r \right\}$ that has value closest to $f_J(\vec{x}_j)$ as the final output.

\subsection{Evaluation}\label{evalsubsect}

Lastly, to evaluate the performance of the final output $f(\vec{x}_j),\ \vec{x}_j \in\PP,$ we use the actual reference values $y_j$ to place $f(\vec{x}_j)$ in the PRED-EGA grid.\\

We repeat the process described in Subsections \ref{layer1subsect} -- \ref{evalsubsect} for a fixed number of trials, after which we report the average of the PRED-EGA grid placements, over all $\vec{x} \in \PP$ and over all trials, as the final evaluation.\\

\begin{figure}
	\begin{center}
		\begin{tikzpicture}[node distance=4cm, every text node part/.style={align=center}]
		\node (input) [blank] {$\vec{x}_j$};
		
		\node (first) [block, right of=input] {$f_o(\vec{x}_j) = S(\CC_{o}^{\star},\vec{x}_j)$ \\
			$f_e(\vec{x}_j) = S(\CC_{e}^{\star},\vec{x}_j)$ \\
			$f_r(\vec{x}_j) = S(\CC_{r}^{\star},\vec{x}_j)$ \\
			$f_J(\vec{x}_j) = S(\CC^{\star},\vec{x}_j)$};
		
		\draw [arrow] (input) -- (first);
		
		\node (second) [block, right of=first, node distance=5cm] {PRED-EGA};
		
		\draw [arrow] (first) -- (second);
		
		\node (output) [blank, right of=second] {$f(\vec{x}_j)$};
		
		\draw [arrow] (second) -- (output);
		
		\end{tikzpicture}
	\end{center}
	\caption{Flowchart diagram: Deep Network for BG prediction}
	\label{diag-1}
\end{figure}
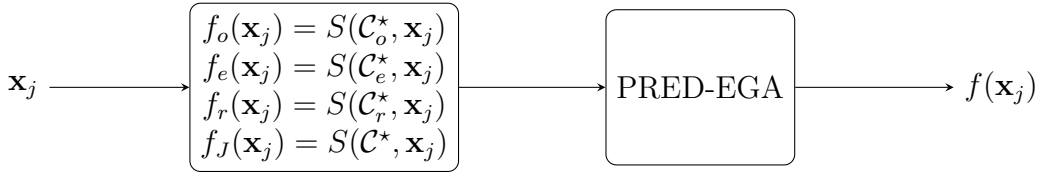

A summary of the method is given in Algorithm \ref{alg}. A flowchart diagram of the algorithm is shown in Figure \ref{diag-1}.

\begin{algorithm}
	\SetKwInOut{Input}{input}
	\SetKwInOut{Output}{output}
	\Input{Time series $\left\{ {s}_p(t_j) \right\}$, $p \in P$, formatted as $\PP = \left\{ \vec{x}_j \right\}$ with \\$\vec{x}_j = \left( s_p(t_{j-d+1}), \cdots, s_p(t_j) \right)$ and $y_j = s_p(t_{j+m})$, and corresponding diffusion matrix\\
		$d \in \N$ (specifies sampling horizon), $m \in \N$ (specifies prediction horizon)\\ $M \in (0,100)$ (percentage of data used for training).\\
		Let $TP$ contain $M\%$ of patients from $P$ (drawn according to uniform prob. distr.) \; \\
		Set $\CC = \left\{ \vec{x}_j \right\}$ and $\CC^{\star} = \left\{ (\vec{x}_j, y_j) \right\}$ for all patients $p \in TP$ \; }
	\Output{Prediction $f(\vec{x}_j)\approx s_p(t_{j+m})$ for $\vec{x}_j\in \mathcal{P}$. }
	\BlankLine

	\SetKwProg{myproc}{}{}{}
	
	\myproc{First layer:}{
		
		\For {$\vec{x}_j \in \CC$} {Make 5-minute prediction $L_{\vec{x}_j}(t_{j+1})$} \;
		Set $\CC_o = \left\{ \vec{x}_j \in \CC: 0 \leq L_{\vec{x}_j}(t_{j+1}) \leq 70 \right\} $ \; \\
		Set $\CC_e = \left\{ \vec{x}_j \in \CC: 70 < L_{\vec{x}_j}(t_{j+1}) \leq 180 \right\} $ \; \\
		Set $\CC_r = \left\{ \vec{x}_j \in \CC: 180 < L_{\vec{x}_j}(t_{j+1}) \leq 450 \right\}$ \; \\
		Set $\CC_{\ell}^{\star} = \left\{ (\vec{x}_j, y_j) : \vec{x}_j \in \CC_{\ell} \right\}, \ \ell \in \left\{ o,e,r\right\}$ \; \\
		\For {$\vec{x}_j \in \PP$} {\For {$\ell \in \left\{ o,e,r \right\}$} {Compute $f_{\ell}(\vec{x}_j) = S(\CC_{\ell}^{\star},\vec{x}_j)$} \;
			Compute $f_J(\vec{x}_j) = S(\CC^{\star}, \vec{x}_j)$ \;
		}  \;
	}
	
	\myproc{Second layer:}{
		\For {$\vec{x}_j \in \PP$} {\For {$\ell \in \left\{ o,e,r \right\}$} {Construct PRED-EGA grid for $f_{\ell}(\vec{x}_j)$ using $f_J(\vec{x}_j)$ as reference value} \;
			Let $\ell_j \in \left\{ o,e,r \right\}$ denote the subscript for which $f_{\ell_j}(\vec{x}_j)$ produces the best PRED-EGA grid placement \; \\
			Set final output $f(\vec{x}_j) = f_{\ell_j}(\vec{x}_j)$
		}
	}

	\caption{Deep Network for BG prediction}
	\label{alg}
\end{algorithm}

\subsection{Remarks}\label{remarksubsect}

\begin{enumerate}[(i)]
	\item An optional smoothing step may be applied before the first training layer step in Subsection \ref{layer1subsect} to remove any large spikes in the given time series $\left\{ s_p(t_j) \right\}, \ p\in P,$ that may be caused by patient movement, for example. Following ideas in \cite{sparacino2007glucose}, we may apply flat low-pass filtering (for example, a first-order Butterworth filter). In this case, the evaluation of the final output in Subsection \ref{evalsubsect} is done by using the original, unsmoothed measurements as reference values.
	
	\item We remark that, as explained in Section \ref{introsect}, the training set may also be implemented in a different way: instead of drawing $M$\% of the patients in $P$ and using their entire data sets for training, we could construct the training set $\CC^{\star}$ for each patient $p \in P$ separately by drawing $M$\% of a given patient's data for training, and then construct the networks $f_{\ell}, \ \ell\in \left\{o,e,r\right\}$ and $f_J$ for each patient separately. This is a different problem, studied often in the literature (see, for example, \cite{reifman2007predictive,sparacino2007glucose,eren2009estimation}). We do not pursue this approach in the current paper.\\
	
\end{enumerate}

\section{Results and discussion}\label{resultsect}

As mentioned in Section \ref{problemsect}, we apply our method to data provided by the DirecNet Central Laboratory. Time series for 25 patients are considered. This specific data set was designed to study the performance of CGM devices in children with Type I diabetes; as such, all of the patients are less than 18 years old. Each time series $\left\{s_p(t_j)\right\}$ contains more than 160 BG measurements, taken at 5-minute intervals. Our method is a
general purpose algorithm, where these details do not play any
significant role, except in affecting the outcome of the experiments.\\

We provide results obtained by implementing our method in Matlab, as described in Algorithm \ref{alg}. For our implementation, we employ a sampling horizon $t_{j}-t_{j-d+1}$ of 30 minutes ($d=7$),  a prediction horizon $t_{j+m} - t_j$ of 30 minutes ($m=6$), and a total of 100 trials ($T=100$). We provide results for both 30\% training data ($M=30$) and 50\% training data ($M=50$) (which is comparable to approaches followed in for example \cite{pappada2011neural,reifman2007predictive}). After testing our method on all 25 patients, the average PRED-EGA scores (in percent) are displayed in Table \ref{table-A}. For 30\% training data, the percentage of accurate predictions and predictions with benign consequences is 84.32\% in the hypoglycemic range, 97.63\% in the euglycemic range, and 82.89\% in the hyperglycemic range, while for 50\% training data, we have 93.21\% in the hypoglycemic range, 97.68\% in the euglycemic range, and 86.78\% in the hyperglycemic range.\\

\begin{table}
	\begin{center}
		\begin{tabular}{c| c c c | c c c | ccc}
			\hline
			& \multicolumn{3}{|c|}{Hypoglycemia:} & \multicolumn{3}{|c|}{Euglycemia:} & \multicolumn{3}{|c}{Hyperglycemia:} \\
			& \multicolumn{3}{|c|}{BG $\leq 70$ (mg/dL)} & \multicolumn{3}{|c|}{BG $70 - 180$ (mg/dL)} & \multicolumn{3}{|c}{BG $> 180$ (mg/dL)} \\
			& Acc. & Benign & Error & Acc. & Benign & Error & Acc. & Benign & Error\\
			\hline
			\multicolumn{10}{l}{\textbf{30\% training data ($M=30$):}} \\ \hline
			Deep network & 79.97 &	4.35 &	15.68 &	81.88 &	15.75 &	2.37 &	62.72 &	20.17 &	17.11 \\ \hline
			Shallow network & 52.79 &	2.64 &	44.57 &	80.55 &	14.04 &	5.41 &	59.37 &	22.09 &	18.54 \\ \hline
			Tikhonov reg. & 52.34 &	2.10 &	45.56 &	81.25 &	13.68 &	5.07 &	61.33 &	19.69 &	18.98 \\ \hline
			\multicolumn{10}{l}{\textbf{50\% training data ($M=50$):}} \\ \hline
			Deep network & 88.72	&4.49&	6.79&	80.32&	17.36 &	2.32&	64.88&	21.90&	13.22\\ \hline
			Shallow network & 51.84&	2.47&	45.69&	80.94&	13.77&	5.29&	60.41&	21.58&	18.01\\ \hline
			Tikhonov reg. & 52.92&	1.70&	45.38&	81.28&	13.66&	5.06&	62.22&	20.26&	17.52 \\ \hline
		\end{tabular}
	\end{center}
	\caption{Average PRED-EGA scores (in percent): $M$\% of patients used for training.}
	\label{table-A}
\end{table}

\begin{figure}
	\begin{center}
		\includegraphics[width=\textwidth]{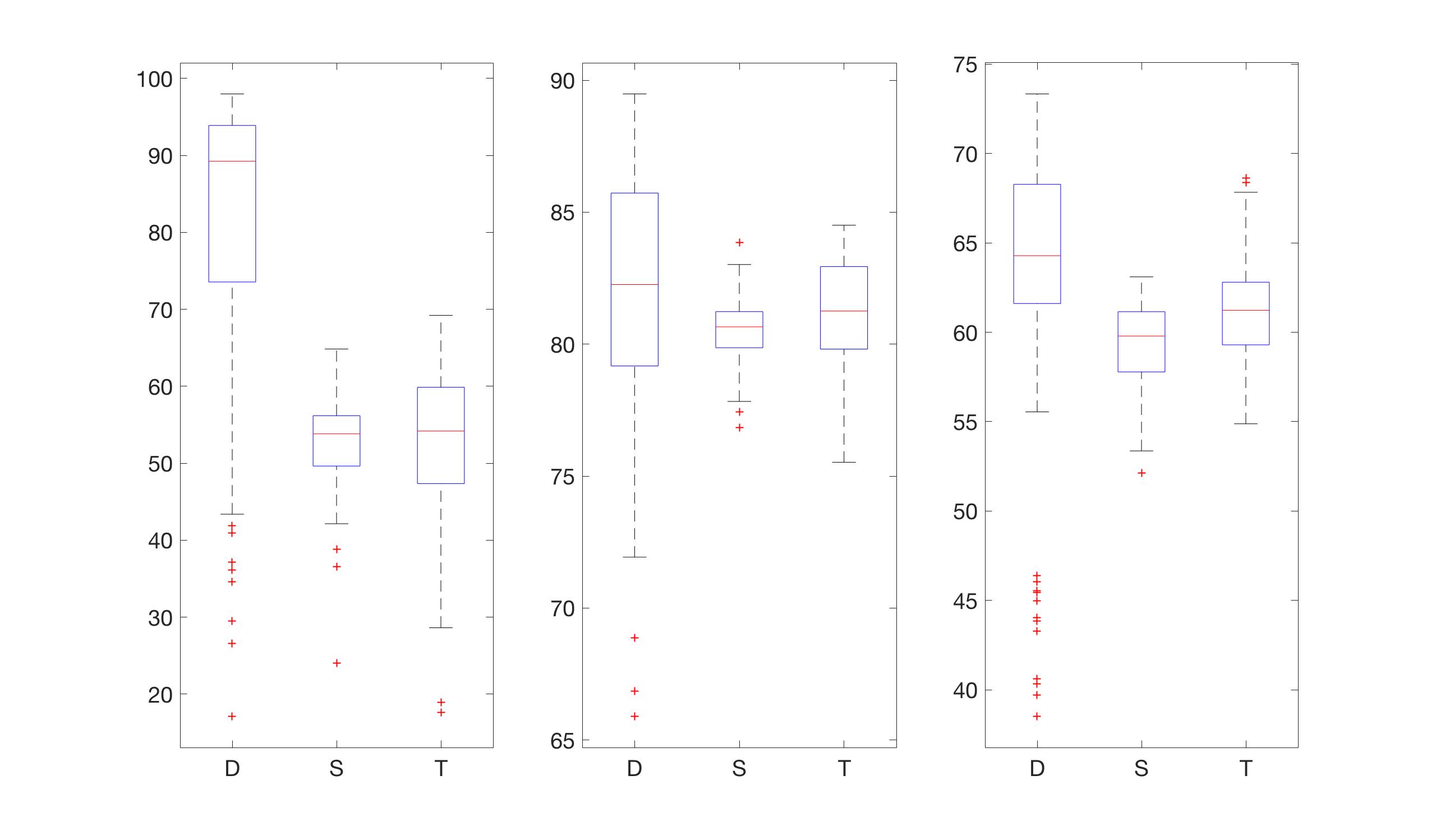}
	\end{center}
	\caption{Boxplot for the 100 experiments conducted with no smoothing and 30\% training data for each prediction method (D = deep network, S = shallow network, T = Tikhonov regularization). The three graphs show the percentage accurate predictions in the hypoglycemic range (left), euglycemic range (middle) and hyperglycemic range (right).}
	\label{fig-1}
\end{figure}

\begin{figure}
	\begin{center}
		\includegraphics[width=\textwidth]{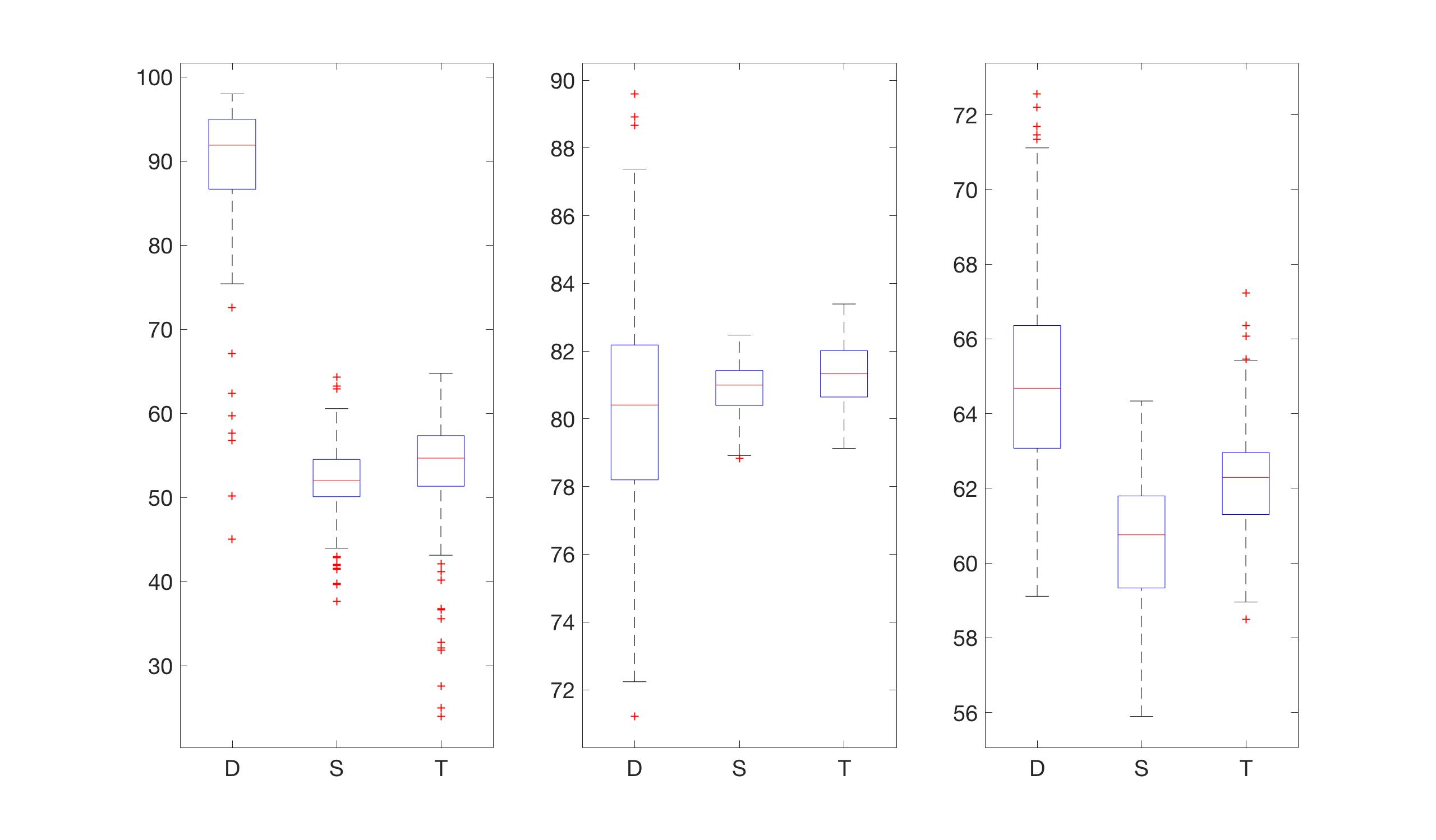}
	\end{center}
	\caption{Boxplot for the 100 experiments conducted with no smoothing and 50\% training data for each prediction method (D = deep network, S = shallow network, T = Tikhonov regularization). The three graphs show the percentage accurate predictions in the hypoglycemic range (left), euglycemic range (middle) and hyperglycemic range (right).}
	\label{fig-2}
\end{figure}

For comparison, Table \ref{table-A} also displays the PRED-EGA scores obtained when implementing a shallow (ie, one layer) feed-forward network with 20 neurons using Matlab's Neural Network Toolbox, using the same parameters $d, m, M$ and $T$ as in our implementation, with Matlab’s default hyperbolic tangent sigmoid activation function. The motivation for using 20 neurons is the following. As mentioned in Subsection \ref{layer1subsect}, each layer in our deep network can be viewed as a classical neural network with exactly as many neurons as the number of eigenfunctions used in \eqref{phidef} and \eqref{eq-sigma}. This number is determined to ensure that the system in \eqref{quadrature} is well conditioned; in our experiments, it turned out to be at most 5. Therefore, to compare our two-layer deep network (where the first layer consists of three separate networks) with a classical shallow neural network, we use 20 neurons. It is clear that our deep network performs substantially better than the shallow network; in all three BG ranges, our method produces a lower percentage of erroneous predictions.\\

For a further comparison, we also display in Table \ref{table-A} the PRED-EGA scores obtained when training through supervised learning using a standard Tikhonov regularization to find the best least square fit to the data; specifically, we implemented the method described in \cite[pp 3-4]{poggio2003mathematics} using a Gaussian kernel with $\sigma = 100$ and regularization constant $\gamma = 0.0001$. Again, our method produces superior results, especially in the hypoglycemic range. \\

Figures \ref{fig-1} and \ref{fig-2} display boxplots for the percentage accurate predictions in each BG range for each method over the 100 trials.

\begin{figure}
	\begin{center}
		\includegraphics[width=\textwidth]{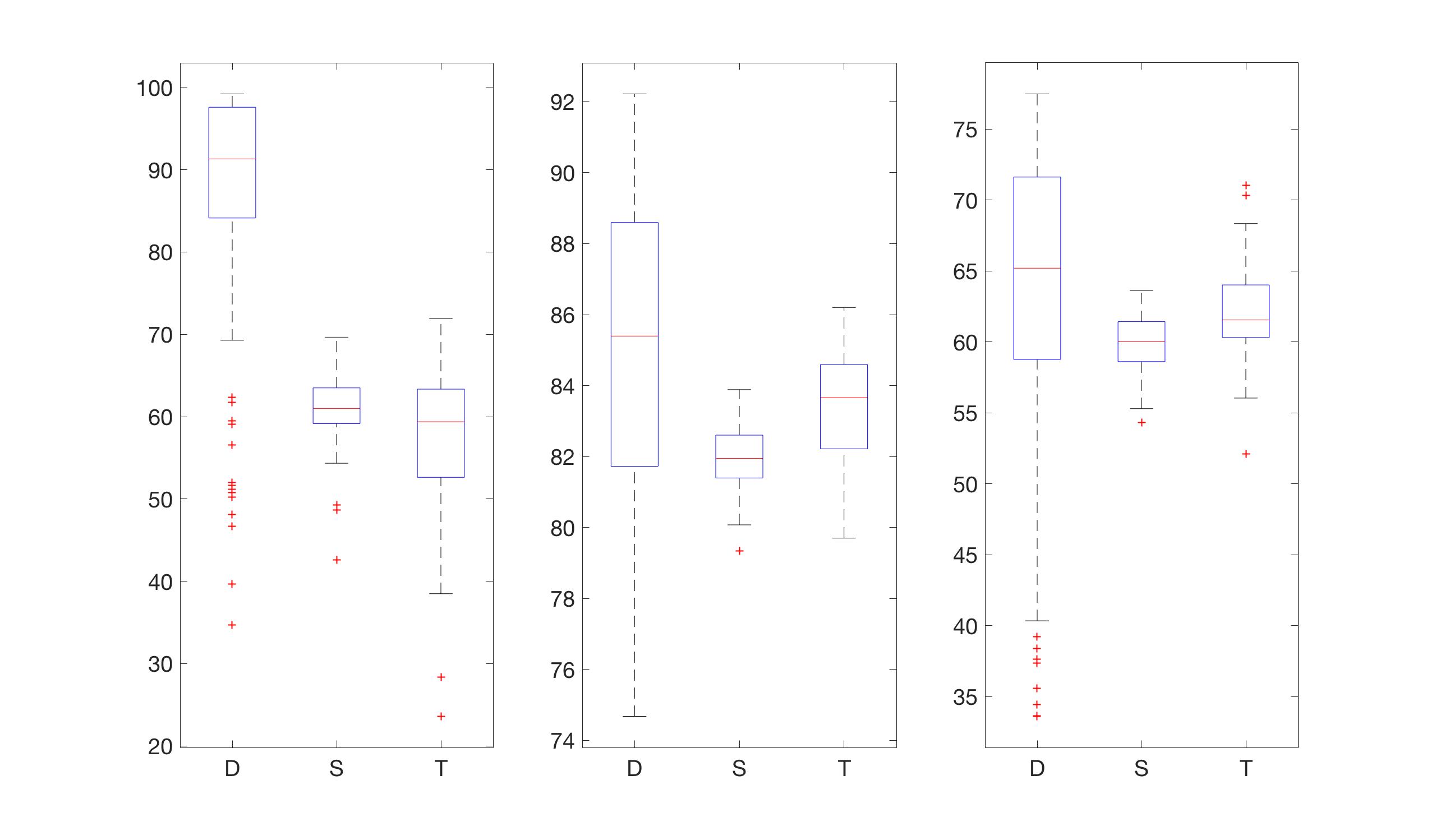}
	\end{center}
	\caption{Boxplot for the 100 experiments conducted with flat low-pass filtering and 30\% training data for each prediction method (D = deep network, S = shallow network, T = Tikhonov regularization). The three graphs show the percentage accurate predictions in the hypoglycemic range (left), euglycemic range (middle) and hyperglycemic range (right).}
	\label{fig-3}
\end{figure}

\begin{figure}
	\begin{center}
		\includegraphics[width=\textwidth]{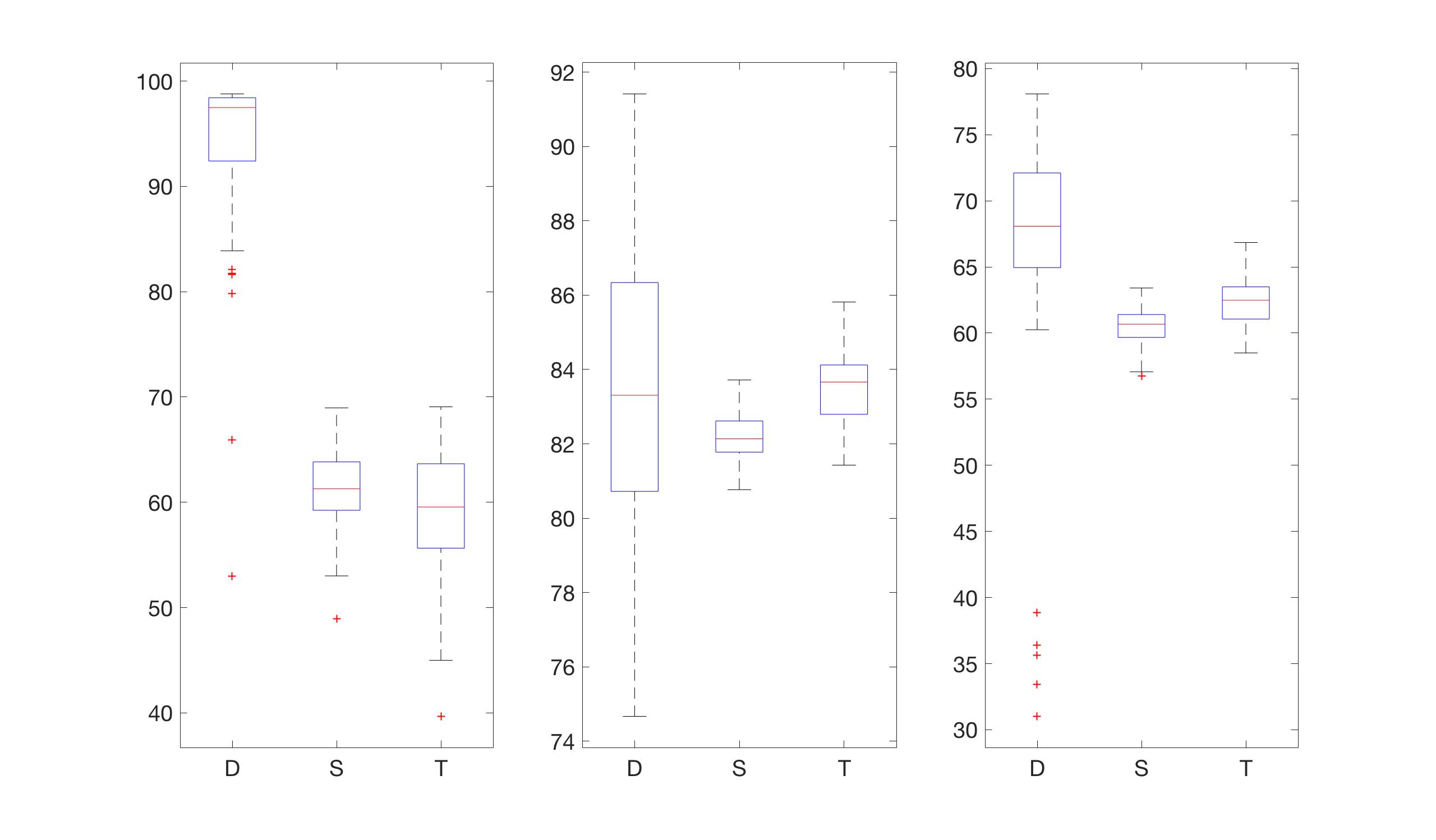}
	\end{center}
	\caption{Boxplot for the 100 experiments conducted with flat low-pass filtering and 50\% training data for each prediction method (D = deep network, S = shallow network, T = Tikhonov regularization). The three graphs show the percentage accurate predictions in the hypoglycemic range (left), euglycemic range (middle) and hyperglycemic range (right).}
	\label{fig-4}
\end{figure}

We also provide results obtained when applying smoothing through flat low-pass filtering to the given time series, as explained in Subsection \ref{remarksubsect}. For our implementations, we used a first-order Butterworth filter with cutoff frequency 0.8, with the same input parameters as before. The results are given in Table \ref{table-C}. For 30\% training data, the percentage of accurate predictions and predictions with benign consequences is 88.92\% in the hypoglycemic range, 97.64\% in the euglycemic range, and 77.58\% in the hyperglycemic range, while for 50\% training data, we have 96.43\% in the hypoglycemic range, 97.96\% in the euglycemic range, and 85.29\% in the hyperglycemic range. Figures \ref{fig-3} and \ref{fig-4} display boxplots for the percentage accurate predictions in each BG range for each method over the 100 trials.\\

\begin{table}
	\begin{center}
		\begin{tabular}{c| c c c | c c c | ccc}
			\hline
			& \multicolumn{3}{|c|}{Hypoglycemia:} & \multicolumn{3}{|c|}{Euglycemia:} & \multicolumn{3}{|c}{Hyperglycemia:} \\
			& \multicolumn{3}{|c|}{BG $\leq 70$ (mg/dL)} & \multicolumn{3}{|c|}{BG $70 - 180$ (mg/dL)} & \multicolumn{3}{|c}{BG $> 180$ (mg/dL)} \\
			& Acc. & Benign & Error & Acc. & Benign & Error & Acc. & Benign & Error\\
			\hline
			\multicolumn{10}{l}{\textbf{30\% training data ($M=30$):}} \\ \hline
			Deep network & 86.41 &	2.51 &	11.08&	85.05&	12.59&	2.36 &	62.24&	15.34&	22.42\\ \hline
			Shallow network & 61.10 &	5.21&	33.69&	81.96 &	12.77&	5.27 &	60.01 &	19.62 &	20.37 \\ \hline
			Tikhonov reg. & 57.47 &	2.01 &	40.52 &	83.49&	12.00&	4.51&	62.13&	19.13&	18.74\\ \hline
			\multicolumn{10}{l}{\textbf{50\% training data ($M=50$):}} \\ \hline
			Deep network & 94.39 &	2.04&	3.57 &	83.44&	14.52&	2.04&	67.21&	18.08&	14.71\\ \hline
			Shallow network & 61.49 &	5.50&	33.01&	82.16&	12.59&	5.25&	60.50&	19.21&	20.29 \\ \hline
			Tikhonov reg. & 59.02&	1.94&	39.04&	83.56&	11.95&	4.49&	62.34&	19.55&	18.11 \\ \hline
		\end{tabular}
	\end{center}
	\caption{Average PRED-EGA scores (in percent): $M$\% of patients used for training with flat low-pass filtering.}
	\label{table-C}
\end{table}

In all our experiments the
percentage of erroneous predictions is substantially smaller with deep
networks than the other two methods we have tried, with the only
exception of hyperglycemic range with 30\% training data and flat low
pass filtering. Our method's performance also improves when the amount of the training data increases from 30\% to 50\%, because the percentage of erroneous predictions decreases in all of the BG ranges in all experiments. \\

Moreover, from this viewpoint, our deep learning method outperforms the considered competitors, except in the hyperglycemic BG range in Table \ref{table-C}. A possible explanation for this is that the DirecNet study was done on children (less than 18 years of age) with type 1 diabetes, for a period of roughly 26 hours. It appears that children usually have prolonged hypoglycemic periods as well as profound postprandial hyperglycemia (high blood sugar “spikes” after meals) – according to \cite{boland2001limitations}, more than 70\% of children display prolonged  hypoglycemia while more than 90\% of children display significant postprandial hyperglycemia. In particular, many of the patients in the data set only exhibit very
limited hyperglycemic BG  spikes – for the patient labeled 43, for
example, there exists a total of 194 BG measurements, of which only 5
measurements are greater than 180 mg/dL. This anomaly might have
affected the performance of our algorithm in the hyperglycemic range.
Obviously, it performs remarkably better than the other techniques we
have tried, including fully supervised training, while ours is only
semi-supervised.

\section{Conclusions}\label{concludesect}
The prediction of blood glucose levels based on continuous glucose monitoring system data 30 minutes ahead is a very important problem with many consequences for the health care industry. 
In this paper, we suggest a deep learning paradigm based on a solid mathematical theory as well as domain knowledge to solve this problem accurately as assessed by the PRED-EGA grid, developed specifically for this purpose.
It is demonstrated in \cite{mhaskar2016deep} that deep networks perform substantially better than shallow networks in terms of expressiveness for function approximation when the target function has a compositional structure. Thus, the blessing of compositionality cures the curse of dimensionality. 
However, the compositional structure is not unique, and it is open problem to decide whether a given target function has a certain compositional structure. 
In this paper, we have demonstrated an example where domain knowledge can be used to build an appropriate compositional structure, leading to a parsimonious deep learning design.

\section*{Acknowledgments}
The research of HNM is supported in part by ARO Grant W911NF-15-1-0385. The research of SVP is partially supported by Austrian Science Fund (FWF)
Grant I 1669-N26.

\bibliographystyle{plain}
\bibliography{bib_bg}


\appendix

\section{Filtered Legendre expansion method}\label{mnpsect}

In this appendix, we review the mathematical background for the method developed in \cite{mhaskar2013filtered} for short term blood glucose prediction. 
As explained in the text, the main mathematical problem can be summarized as that of estimating the derivative of a function at the end point of an interval, based on measurements of the function in the past. 
Mathematically, if $f: [-1,1]\to\mathbb{R}$ is continuously differentiable, we wish to estimate $f'(1)$ given the noisy values $\{y_j=f(t_j)+\epsilon_j\}$ at points $\{t_j\}_{j=1}^d \subset [-1,1]$. We summarize only the method here, and refer the reader to \cite{mhaskar2013filtered} for the detailed proof of the mathematical facts.\\

For this purpose, we define the Legendre polynomials recursively by
\begin{equation}\label{legendre_rec}
P_k(x)=\frac{2k-1}{2k}xP_{k-1}(x)-\frac{k-1}{k}P_{k-2}(x), \ k=2,3,\ldots;\quad P_0(x)=1,\ P_1(x)=x.
\end{equation} 
Let $h : \mathbb{R}\to\mathbb{R}$ be an even, infinitely differentiable function, such that $h(t)=1$ if $t\in [0,1/2]$ and $h(t)=0$ if $t\ge 1$. We define
\begin{equation}\label{diffkerndef}
K_n(h;x)=\frac{1}{2}\sum_{k=0}^{n-1} h\left(\frac{k}{n}\right)k(k+1/2)(k+1)P_k(x).
\end{equation}

Next, given the points $\{t_j\}_{j=1}^d\subset [-1,1]$, we use least squares to find $n=n_d$ such that 
\begin{equation}
\sum_{j=1}^n w_jP_k(t_j)=\int_{-1}^1 P_k(t)dt=\left\{\begin{array}{ll}
2, &\mbox{ if $k=0$},\\
0, & \mbox{ if $1\le k <2n$},
\end{array}\right.
\label{intervalquad}
\end{equation}
and the following estimates are valid for all polynomials of degree $<2n$:
\begin{equation}
\sum_{j=1}^n |w_jP(t_j)|\le A\int_{-1}^1 |P(t)|dt,
\label{intervalmz}
\end{equation}
for some positive constant $A$. We do not need to determine $A$, but it is guaranteed to be proportional to the condition number of the system in \eqref{intervalquad}. \\

Finally, our estimate of $f'(1)$ based on the values $y_j=f(t_j)+\epsilon_j$, $j=1,\cdots, d$, is given by
\begin{equation}\label{intervalderop}
\mathcal{S}_n(h;\{y_j\})=\sum_{j=1}^d w_j y_jK_n(h;t_j).
\end{equation}

It is proved in \cite[Theorem~3.2]{mhaskar2013filtered} that if $f$ is twice continuously differentiable, 
$$
\Delta_{[-1,1]}(f)(x)=2xf'(x)-(1-x^2)f''(x),
$$
and
$|\epsilon_j|\le \delta$ then
\begin{equation}\label{intervalderest}
|\mathcal{S}_n(h;\{y_j\})-f'(1)|\le cA\left\{E_{n/2,[-1,1]}(\Delta_{[-1,1]}(f))+n^2\delta\right\},
\end{equation}
where
\begin{equation}
E_{n/2,[-1,1]}(\Delta_{[-1,1]}(f))=\min\max_{x\in [-1,1]}|\Delta_{[-1,1]}(f)(x)-P(x)|,
\label{intervaldegapprox}
\end{equation}
the minimum being over all polynomials $P$ of degree $<n/2$. 

\section{Function approximation on data defined spaces}\label{funcapproxsect}

While classical approximation theory literature deals with function approximation based on data that is dense on a known domain, such as a cube, torus, sphere, etc.,
this condition is generally not satisfied in the context of machine learning. 
For example,  the set of vectors $(s_p(t_j),\ldots, s_p(t_j-d+1))$ is unlikely to be dense in a cube in $\mathbb{R}^d$. 
A relatively recent idea is to think of the data as being sampled from a distribution on an unknown manifold, or more generally, a locally compact, quasi-metric measure space. 
In this appendix, we review the theoretical background that underlies our experiments reported in this paper. The discussion here is based mainly on \cite{maggioni2008diffusion,mhaskar2010eignets,ehler2012locally}.\\

Let $\mathbb{X}$ be a locally compact quasi-metric measure space, with $\rho$ being the quasi-metric and $\mu^*$ being the measure. 
In the context of machine learning, $\mu^*$ is a probability measure and $\mathbb{X}$ is its support. 
The starting point of this theory is a non-decreasing sequence of numbers $\{\lambda_k\}_{k=0}^\infty$, such that $\lambda_0=0$ and $\lambda_k\to \infty$ as $k\to\infty$, and a corresponding sequence of bounded functions $\{\phi_k\}_{k=0}^\infty$ that forms an orthonormal sequence in $L^2(\mu^*)$. 
Let $\Pi_\lambda=\mathsf{span}\{\phi_k : \lambda_k <\lambda\}$, and analogously to \eqref{intervaldegapprox}, we define for a uniformly continuous, bounded function $f: \mathbb{X}\to\mathbb{R}$,
\begin{equation}\label{degapproxdef}
E_\lambda(f)=\min_{P\in\Pi_\lambda}\sup_{x\in\mathbb{X}}|f(x)-P(x)|.
\end{equation}
With the function $h$ as defined in Appendix~\ref{mnpsect}, we define
\begin{equation}\label{phidef}
\Phi_\lambda(h;x,y)=\sum_{k : \lambda_k <\lambda}h\left(\frac{\lambda_k}{\lambda}\right)\phi_k(x)\phi_k(y), \quad x,y \in \mathbb{X}.
\end{equation}

Next,  let $\{x_j\}_{j=1}^M\subset \mathbb{X}$ be the ``training data''. Our goal is to learn a function $f :\mathbb{X}\to\mathbb{R}$ based on the values $\{f(x_j)\}_{j=1}^M$. Toward this goal, we solve an under-determined system of equations
\begin{equation}\label{quadrature}
\sum_{j=1}^M W_j \phi_k(x_j)=\left\{\begin{array}{ll}
1, &\mbox{ if $k=0$},\\
0, &\mbox{ if $k>0$},
\end{array}\right. \qquad k : \lambda_k<\lambda
\end{equation}
for the largest possible $\lambda$ for which this system is still well conditioned. We then define an approximation, analogous to the classical radial basis function networks, by
\begin{equation}
\label{eq-sigma}
\sigma_\lambda(h;f,x)=\sum_{j=1}^M W_j f(x_j)\Phi_\lambda(x,x_j), \quad x \in \mathbb{X}.
\end{equation}

It is proved in \cite{maggioni2008diffusion,mhaskar2010eignets,ehler2012locally} that under certain technical conditions, for a uniformly continuous, bounded function $f: \mathbb{X}\to\mathbb{R}$, we have 
\begin{equation}\label{goodapprox}
E_\lambda(f)\le \sup_{x\in\mathbb{X}}|f(x)-\sigma_\lambda(h;f,x)|\le cE_{\lambda/2}(f),
\end{equation}
where $c>0$ is a generic positive constant. Thus, the approximation $\sigma_\lambda(f)$ is guaranteed to yield a ``good approximation'' that is asymptotically best possible given the training data. \\

In practice, one has a point cloud $\mathcal{P}=\{y_i\}_{i=1}^N \supset \{x_j\}_{j=1}^M$ rather than the space $\mathbb{X}$. 
The set $\mathcal{P}$ is a subset of some Euclidean space $\mathbb{R}^D$ for a possibly high value of $D$, but is assumed to lie on a compact sub-manifold $\mathbb{X}$ of $\mathbb{R}^D$, with this manifold having a low dimension.
We build the graph Laplacian from $\mathcal{P}$ as follows: 
With a parameter $\varepsilon>0$,  induces the weight matrix $\mathcal{W}^\varepsilon_N$ defined by 
\begin{equation}
\mathcal{W}^\varepsilon_{N;i,j} = \exp(-|y_i-y_j|^2/\varepsilon) .
\label{eq-diffusion-matrix}
\end{equation}
We build the diagonal matrix $D^\varepsilon_{N;i,i}=\sum_{j=1}^N \mathcal{W}^\varepsilon_{N;i,j}$ and define the (unnormalized) graph Laplacian as 
\begin{equation}
\label{graphlaplacian}
L^\varepsilon_N=\mathcal{W}^\varepsilon_N-D^\varepsilon_N.
\end{equation}
Various other versions of this Laplacian are also used in practice.
For $M\rightarrow \infty$ and $\varepsilon\rightarrow 0$, the eigenvalues and interpolations of the eigenvectors of $L^\varepsilon_M$ converge towards the ``interesting'' eigenvalues $\lambda_k$ and eigenfunctions $\phi_k$ of a differential operator $\Delta_\mathbb{X}$. This behavior has been studied in detail by many authors, e.g., \cite{belkin2005towards,lafon2004diffusion,singer2006graph}.
When $\{y_i\}_{i=1}^M$ are uniformly distributed on $\mathbb{X}$, this operator is the Laplace-Beltrami operator on $\mathbb{X}$. 
If $\{y_i\}_{i=1}^M$ are distributed according to the density $p$, then the graph Laplacian approximates the elliptic Schr\"odinger type operator $\Delta +\frac{\Delta p}{p}$, whose eigenfunctions $\phi_k$ also form an orthonormal basis for $L_2(\mathbb{X},\mu)$.

\end{document}